\documentclass[letterpaper, 10 pt, conference]{ieeeconf}  % Comment this line out if you need a4paper

\IEEEoverridecommandlockouts                              % This command is only needed if 
\usepackage{graphicx}
\usepackage{alltt}
\usepackage{booktabs}
\usepackage{multirow}
\usepackage{float}

\title{\LARGE \bf
A Preliminary Study on the Learning Informativeness of Data Subsets
}

\author{Simon Kaltenbacher$^{1}$, Nicholas H. Kirk$^{2}$ and Dongheui Lee$^{2}$% <-this % stops a space
\thanks{$^{1}$S.K. is with Ludwig Maximilian University of Munich, Germany
        {\tt\small simon.kaltenbacher@campus.lmu.de}}%
\thanks{$^{2}$N.H.K. and D.L. are with the Technical University of Munich, Germany
        {\tt\small \{nicholas.kirk,dhlee\}@tum.de}}%
}

\begin{document}

\maketitle
\thispagestyle{empty}
\pagestyle{empty}
\noindent Estimating the internal state of a robotic system is complex: this is performed from multiple heterogeneous sensor inputs and knowledge sources.
Discretization of such inputs is done to capture saliences, represented as \textit{symbolic} information, which often presents structure and recurrence.
As these sequences are used to reason over complex scenarios \cite{kirk15predicting}, a more compact representation would aid exactness of technical cognitive reasoning capabilities, which are today constrained by computational complexity issues and fallback to representational heuristics or human intervention \cite{kirk15predicting,kirk2014controlled}. Such problems need to be addressed to ensure timely and meaningful human-robot interaction.

Our work is towards understanding the variability of learning informativeness when training on subsets of a given input dataset. This is in view of reducing the training size while retaining the majority of the symbolic learning potential.
We prove the concept on human-written texts, and conjecture this work will reduce training data size of sequential instructions, while preserving semantic relations, when gathering information from large remote sources \cite{tenorth2010understanding}.

\subsection*{Posterior Evaluation Distribution of Subsets}
\begin{figure}[t!]
\vspace{2mm}
\resizebox{0.99\linewidth}{!}{  
\centerline{%
\includegraphics[scale=0.60]{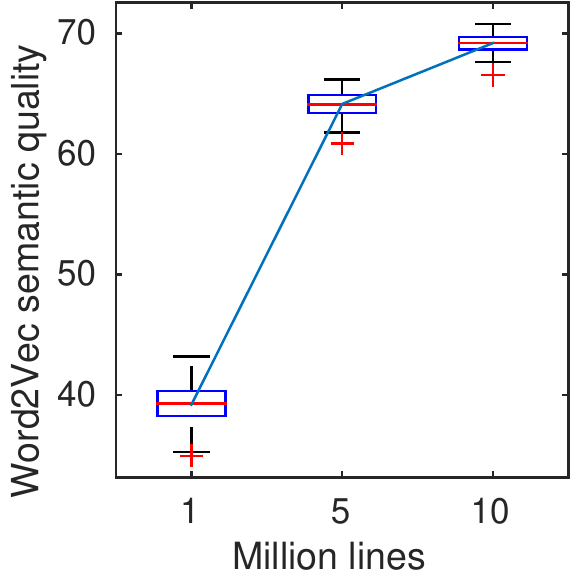}%
\hspace{2mm}
\includegraphics[scale=0.60] {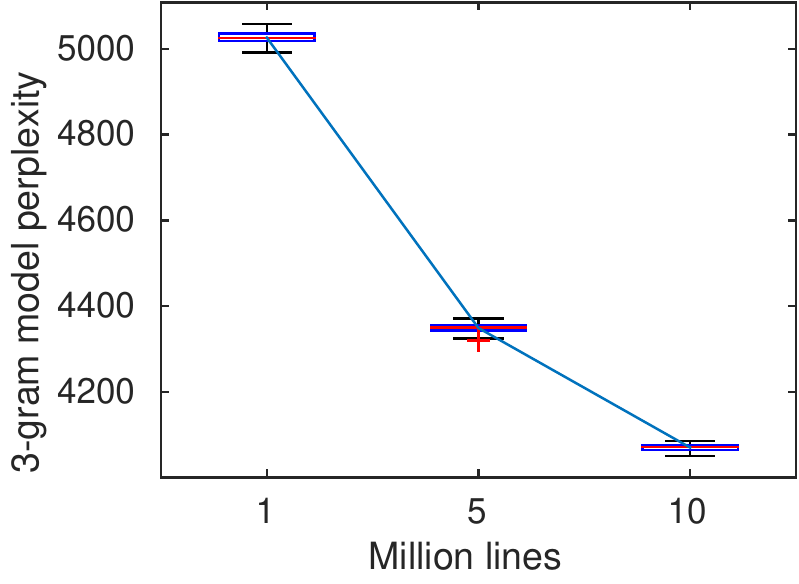}%
}%
}
\caption{Evaluation values for random subselections of various sizes, for both semantic and syntactic tasks (100 instances for each visualized size).}
\label{fig:normality}
\end{figure}

\noindent We computed multiple random subsets of sentences from the \textsc{UMBC Webbase corpus} %\cite{han2013umbc} 
($\sim 17.13$GB) via a custom implementation using the \textsc{Spark} distributed framework. %\cite{zaharia2012resilient}. 
We evaluated the learning informativess of such sets in terms of semantic word-sense classification accuracy (with \textsc{Word2Vec} \cite{mikolov2013distributed}), and of n-gram perplexity.
Previous literature inform us that corpus size and posterior quality do not follow linear correlation for some learning tasks (e.g. semantic measures) \cite{banko2001scaling}. In our semantic tests, on average $85\%$ of the quality can be obtained by training on a random $\sim 4\%$ subset of the original corpus (e.g. as in Fig. \ref{fig:normality}, 5 random million lines yield $64.14\%$ instead of $75.14\%$). 

Our claims are that i) such evaluation posteriors are Normally distributed (Tab. \ref{tab:nonuniform}), and that ii) the variance is inversely proportional to the subset size (Tab. \ref{tab:variance}).

\noindent It is therefore possible to select the best random subset for a given size, if an information criterion is known. Such metric is currently under investigation.
Within the robotics domain, in order to reduce computational complexity of the training phase, cardinality reduction of human-written instructions is particularly important for non-recursive online training algorithms, such as current symbol-based probabilistic reasoning systems \cite{kirk15predicting,tenorth2010understanding,kirk2014towards}.

\begin{table}[t!]

\vspace{2mm}
\centering
\scriptsize

\scalebox{0.81}{
\begin{tabular}{lrcccccccc} \toprule
& & \multicolumn{2}{c}{100 subsets of 1M} & \multicolumn{2}{c}{100 subsets of 5M} & \multicolumn{2}{c}{100 subsets of 10M}\\
& &  $h$ &  $p$ &  $h$ & $p$ &  $h$ & $p$\\ 
\cmidrule(r){3-4} \cmidrule(l){5-6} \cmidrule(l){7-8}
\multirow{2}{*}{\textsc{Word2vec}} & $\chi^2$ & 0 & 0.4221 & 0 & 0.5756 & 0 & 0.9189 \\
& And.-Darling & 0 & 0.8749 & 0 & 0.7616 & 0 & 0.8710\\
\multirow{2}{*}{\textsc{Perplexity}} & $\chi^2$ & 0 & 0.2963 & 0 & 0.2435 & 0 & 0.2443 \\
& And.-Darling & 0 & 0.4908 & 0 & 0.1488 & 0 & 0.3423\\
\bottomrule
\end{tabular}
}
\caption{\scriptsize Chi-square and Anderson-Darling tests showing there is no Gaussian null hypothesis rejection for word2vec and perplexity accuracy values of random subsets (10\% significance level).}
\label{tab:nonuniform}
\end{table}

\begin{table}[t!]
\centering
\scriptsize
\vspace{-2mm} 
\scalebox{0.97}{
\begin{tabular}{lccc} \toprule
&  \multicolumn{1}{c}{100 subsets of 1M} & \multicolumn{1}{c}{100 subsets of 5M} & \multicolumn{1}{c}{100 subsets of 10M}\\
&  \multicolumn{1}{c}{variance} & \multicolumn{1}{c}{variance} & \multicolumn{1}{c}{variance}\\ 
\cmidrule(r){2-2} \cmidrule(l){3-3} \cmidrule(l){4-4}
\multirow{1}{*}{\textsc{Word2vec}} & 2.6199  & 1.0351 & 0.6147 \\
\multirow{1}{*}{\textsc{Perplexity}}& 213.21 & 118.87 & 55.218 \\
\bottomrule
\end{tabular}
}
\caption{\scriptsize Variance values of word2vec and perplexity accuracy posteriors of random subsets.}
\label{tab:variance}
\vspace{-6mm}
\end{table}

\bibliographystyle{ieeetr}
\bibliography{hfr.bib}

\end{document}